\documentclass{article}
\usepackage{graphicx}
\usepackage{authblk}
\usepackage{hyperref}

\title{Solving Even-Parity Problems using \\
Traceless Genetic Programming}

\author{Mihai Oltean}
\affil{Department of Computer Science\\
Faculty of Mathematics and Computer Science\\
Babe\c s-Bolyai University, Kog\u alniceanu 1\\
Cluj-Napoca, 3400, Romania\\
Email: mihai.oltean@gmail.com}

\date{2004.01.01}

\begin{document}
\maketitle

\begin{abstract}

A new Genetic Programming (GP) variant called Traceless Genetic Programming (TGP) is proposed in this paper. TGP is a hybrid method combining a technique for building the individuals and a technique for representing the individuals.
The main difference between TGP and other GP techniques is that TGP does not explicitly store the evolved computer programs. Two genetic operators are used in conjunction with TGP: crossover and insertion. TGP is applied for evolving digital circuits for the even-parity problem. Numerical experiments show that TGP outperforms standard GP with several orders of magnitude.

\end{abstract}

\section{Introduction}

Traceless Genetic Programming (TGP) \footnote {The source code for TGP is available at \url{https://github.com/mihaioltean/Traceless-Genetic-Programming}} is a GP \cite{koza1,koza2} variant as it evolves a population of computer programs. TGP is a hybrid method combining a technique for building the individuals and technique for representing the individuals.

The main difference between the TGP and GP is that TGP does not explicitly store the evolved computer programs. TGP is useful when the trace (the way in which the results are obtained) between the input and output is not important. In this way the space used by traditional techniques for storing the entire computer programs (or mathematical expressions in the simple case of symbolic regression) is saved.

We choose to apply the proposed TGP technique to the even-parity problems because according to Koza \cite{koza1} these problems appear to be the most difficult Boolean functions to be detected via a blind random search.

Evolutionary techniques have been extensively used for evolving digital circuits \cite{coello1,koza1,koza2,oltean1,miller1,miller2,miller3,miller4,stoica1}, due to their practical importance. The case of even-parity circuits was deeply analyzed \cite{koza1,koza2,oltean1,miller4} due to their simple representation.

Special techniques have been proposed in order to improve in standard GP: Automatically Defined Functions \cite{koza2}, Sub-machine code GP \cite{poli1} and Sub-symbolic node representation \cite{poli2}. 

Standard GP was able to solve up to even-5 parity. Using the proposed TGP we are able to easily solve up to even-8 parity problem. Numerical experiments show that TGP outperforms standard GP with several order of magnitude. 

The paper is organized as follows: In section \ref{tgp} the Traceless Genetic Programming technique is described. The parity problem is briefly presented in section \ref{parity}. In section \ref{exp} several numerical experiments for solving the parity problems are performed.

\section{Traceless Genetic Programming}\label{tgp}

In this section the proposed TGP technique is described. TGP is a hybrid method combining a technique for building the individuals and a technique for representing the individuals.

\subsection{Prerequisite}

The quality of a GP individual is usually computed using a set of fitness cases \cite{koza1,koza2}. For instance, the aim of symbolic regression is to find a mathematical expression that satisfies a set of $m$ fitness cases. 

We consider a problem with $n$ inputs: $x_{1}$, $x_{2}$, \ldots $x_{n}$ and one output $f$. The inputs are also called terminals \cite{koza1}. The function symbols that we use for constructing a mathematical expression are $F=\{+,-,*,/, sin\}$.

Each fitness case is given as a ($n+1$) dimensional array of real values:\\

\[
v_1^k ,v_2^k ,v_3^k ,...,v_n^k ,f_k 
\]

\noindent
where $v_j^k $ is the value of the $j^{th}$ attribute (which is $x_{j})$ in 
the $k^{th}$ fitness case and $f_{k}$ is the output for the $k^{th}$ fitness 
case.

Usually more fitness cases are given (denoted by $m$) and the task is to find the expression that best satisfies all these fitness cases. This is usually done by minimizing the quantity:

\[
Q=\sum\limits_{k = 1}^m {\left| {f_k - o_k } \right|} ,
\]

\noindent
where $f_{k}$ is the target value for the $k^{th}$ fitness case and $o_{k}$ is 
the actual (obtained) value for the $k^{th}$ fitness case.

\subsection{Individual representation}

Each TGP individual represents a mathematical expression evolved so far, but 
the TGP individual does not explicitly store this expression. Each TGP 
individual stores only the obtained value, so far, for each fitness case. Thus 
a TGP individual is:\\

($o_{1}$, $o_{2}$, $o_{3}$, \ldots , $o_{m})^{T}$,\\

\noindent
where $o_{k}$ is the current value for the $k^{th}$ fitness case. Each 
position in this array (a value $o_{k})$ is a gene. As we said it before 
behind these values is a mathematical expression whose evaluation has 
generated these values. However, we do not store this expression. We store 
only the values $o_{k}$.\\

\textbf{Remark}

The structure of an TGP individual can be easily enhanced for storing the evolved computer program (mathematical expression). Storing the evolved expression can provide a more easy way to analyze the results of the numerical experiments. However, in this paper, we do not store the trees associated with the TGP individuals.

\subsection{Initial population}

The initial population contains individuals whose values have been generated 
by simple expressions (made up a single terminal). For instance, if an 
individual in the initial population represent the expression:\\

$E=x_{1}$,\\

\noindent
then the corresponding TGP individual is represented as:\\

\[
C = (v_1^1 ,v_1^2 ,v_1^3 ,...,v_1^m)
\]

\noindent
where $v_j^k $ has been previously explained.

The quality of this individual is computed using the equation previously 
described:

\[
Q = \sum\limits_{i = 1}^m {\left| {v_1^k - f_k } \right|} .
\]

\subsection{Genetic Operators}

In this section the genetic operators used in conjunction with TGP are 
described. TGP uses two genetic operators: crossover and insertion. These operators are specially designed for the proposed TGP technique.

\subsubsection{Crossover}

The crossover is the only variation operator that creates new individuals. For 
crossover several individuals (the parents) and a function symbol are 
selected. The offspring is obtained by applying the selected 
operator for each of the genes of the parents.

\begin{figure*}[htbp]
\centerline{\includegraphics{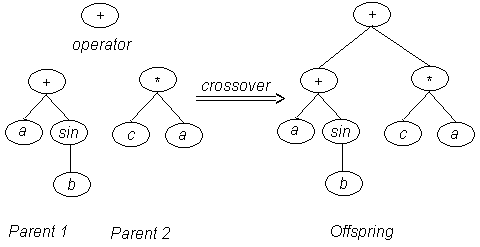}}
\caption{An example of TGP crossover.}
\label{tgp_cross}
\end{figure*}

Speaking in terms of expressions, an example of TGP crossover is depicted in Figure \ref{tgp_cross}.

From Figure \ref{tgp_cross} we can see that the parents are subtrees of the offspring.

The number of parents selected for crossover depends on the number of 
arguments required by the selected function symbol. Two parents have to be selected for crossover if the function symbol is a binary operator. A single parent needs to be selected if the function symbol is a unary operator.\\

\textbf{Example 1}\\

Let us suppose that the operator + is selected. In this case two parents: \\

$C_{1}$ = ($p_{1}$, $p_{2}$, \ldots , $p_{m})^{T}$ and

$C_{2}$ = ($q_{1}$, $q_{2}$, \ldots , $q_{m})^{T}$\\

\noindent
are selected and the offspring $O$ is obtained as follows:\\

$O$ = ($p_{1}+q_{1}$, $p_{2}+q_{2}$,\ldots , $p_{m}+q_{m})^{T}$.\\

\textbf{Example 2}\\

Let us suppose that the operator \textit{sin} is selected. In this case one parent:\\

$C_{1}$ = ($p_{1}$, $p_{2}$, \ldots , $p_{m})^{T}$ \\

\noindent
is selected and the offspring $O$ is obtained as follows:\\

$O$ = (\textit{sin}($p_{1})$, \textit{sin}($p_{2})$,\ldots , \textit{sin}($p_{m}))^{T}$.\\

\subsubsection{Insertion}

This operator inserts a simple expression (made up of a single terminal) in the population. This operator is useful when the population contains individuals representing very complex expressions that cannot improve the search. By inserting simple expressions we give a chance to the evolutionary process to choose another direction for evolution.

\subsection{TGP Algorithm}

Due to the special representation and due to the newly proposed genetic 
operators, TGP uses a special generational algorithm which is given below:

The TGP algorithm starts by creating a random population of individuals. The evolutionary process is run for a fixed number of generation. At each generation the following steps 
are repeated until the new population is filled: With a probability $p_{insert}$ generate an offspring made up of a single terminal (see the Insertion operator). With a probability 1-$p_{insert}$ select two parents using a standard selection procedure. The parents are recombined in order to obtain an offspring. 
The offspring enters the population of the next generation. 

\begin{figure*}[htbp]
\centerline{\includegraphics{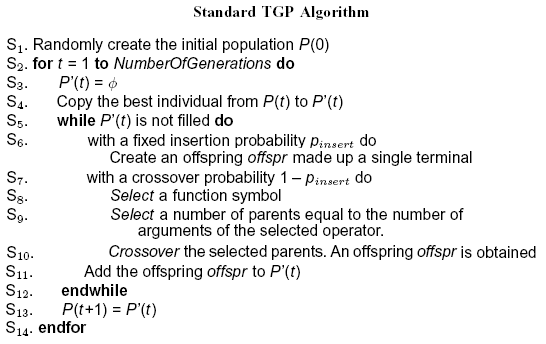}}
\caption{Traceless Genetic Programming Algorithm.}
\label{tgp_alg}
\end{figure*}

The standard TGP algorithm is depicted in Figure \ref{tgp_alg}.

\subsection{Complexity of the TGP Decoding Process} \label{complexity}

A very important aspect of the GP techniques is the time complexity of the procedure used for computing the fitness of the newly created individuals.

The complexity of that procedure for the standard GP is: \\

$O(m*g)$,\\

where $m$ is the number of fitness cases and $g$ is average number of nodes in the GP tree.

By contrast, the TGP complexity is only \\

$O(m)$ \\

because the quality of a TGP individual can be computed by traversing it only once. The length of a TGP individual is $m$.

Due to this reason we may allow TGP programs to run $g$ times more generations in order to obtain the same complexity as the standard GP.

\section{Parity Problem}\label{parity}

The Boolean even-parity function of $k$ Boolean arguments returns \textbf{T} 
(\textbf{True}) if an even number of its arguments are \textbf{T}. Otherwise 
the even-parity function returns \textbf{NIL} (\textbf{False}) \cite{koza1}.

In applying TGP to the even-parity function of $k$ arguments, the terminal set 
$T$ consists of the $k$ Boolean arguments $d_{0}$, $d_{1}$, $d_{2}$, ... $d_{k - 1}$. 
The function set $F$ consists of four two-argument primitive Boolean functions: 
AND, OR, NAND, NOR. According to \cite{koza1} the Boolean even-parity functions 
appear to be the most difficult Boolean functions to be detected via a blind 
random search.

The set of fitness cases for this problem consists of the 2$^{k}$ combinations 
of the $k$ Boolean arguments. The fitness of an TGP chromosome is the sum, over 
these 2$^{k}$ fitness cases, of the Hamming distance (error) between the 
returned value by the TGP chromosome and the correct value of the Boolean 
function. Since the standardized fitness ranges between 0 and 2$^{k}$, a 
value closer to zero is better (since the fitness is to be minimized).

Several techniques have been used in the past for solving the parity problems \cite{koza1,koza2,oltean1}.

\section{Numerical Experiments}\label{exp}

Several numerical experiments using TGP are performed in this section using the even-parity problems. General parameter of the TGP algorithm are given in Table \ref{tab1}.

\begin{table}[htbp]
\caption{General parameters of the TGP algorithm for solving parity problems.}
\label{tab1}
\begin{center}
\begin{tabular}
{|p{100pt}|p{100pt}|}
\hline
\textbf{Parameter}& 
\textbf{Value} \\
\hline
Insertion probability& 
0.05 \\
\hline
Selection& 
Binary Tournament \\
\hline
Function set {\{}gates{\}}& 
{\{}AND, OR, NAND, NOR{\}} \\
\hline
Terminal set& 
Problem inputs \\
\hline
Number of runs& 
100 \\
\hline
\end{tabular}
\end{center}
\end{table}

For assessing the performance of the TGP algorithm the computational effort an the probability of success metrics \cite{koza1} are used.

The method used to assess the effectiveness of an algorithm has been suggested by Koza \cite{koza1}. It consists of calculating the number of chromosomes, which would have to be processed to give a certain probability of success. 
To calculate this figure one must first calculate the cumulative probability of success $P(M, i)$, where $M$ represents the population size, and $i$ the generation number. The value $R(z)$ represents the number of independent runs required for a 
probability of success (given by $z)$ at generation $i$. The quantity \textit{I(M, z, i) }represents the minimum number of chromosomes which must be processed to give a probability of success $z$, at generation $i$. The formulae are given by the 
equation (\ref{eq6}), (\ref{eq7}) and (\ref{eq8}). \textit{Ns}($i)$ represents the number of successful runs at generation $i$, and $N_{total}$, represents the total number of runs:

\begin{equation}
\label{eq6}
P(M,i) = \frac{Ns(i)}{N_{total} }.
\end{equation}

\begin{equation}
\label{eq7}
R(z) = ceil\left\{ {\frac{\log (1 - z)}{\log (1 - P(M,i)}} \right\}.
\end{equation}

\begin{equation}
\label{eq8}
I(M,i,z) = M \cdot R(z) \cdot i.
\end{equation}

In the tables and graphs of this section $z$ takes the value 0.99.

\subsection{Even-3 parity}

The number of fitness cases for this problem is $2^{3}=8$. For solving the even-3 parity we use a population of 50 individuals evolved for 200 generations. Other TGP parameters are given in Table \ref{tab1}.

\begin{figure}[htbp]
\centerline{\includegraphics{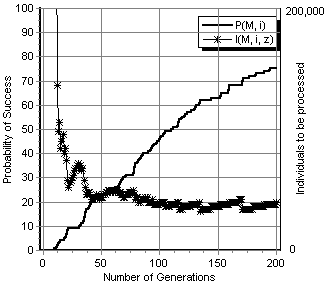}}
\caption{The cumulative probability of success and the computational effort for the even-3 parity problem. Results are averaged over 100 runs.}
\label{fig1}
\end{figure}

The effort and the probability of success of the TGP algorithm are depicted in Figure \ref{fig1}. 

The minimum effort is 33,750 and it was obtained at generation 131.
We want to compare the result obtained by TGP with that obtained by standard GP. 

In \cite{koza1} GP was used for solving the even-3 parity problem using a population of 4000 individuals evolved for 51 generations. The results indicated that 80,000 individuals are sufficient to be processed in order to obtain a solution for this problem \cite{koza1}. One of the obtained solutions is a tree with 45 nodes.

As we noted in section \ref{complexity}, the complexity of computing the fitness of the TGP individuals is $g$ times lower ($g$ is the number of nodes in a GP tree) than the complexity of decoding GP individuals. 

Due to this reason we have to divide the TGP effort (33,750) by 45 (the number of nodes in a GP tree for the even-3 parity problem). Thus, the actual TGP effort is 750 which is with 2 orders of magnitude better than the result obtained by GP. Note that a perfect comparison between GP and TGP cannot be made due to their different individual representation.

\subsection{Even-4 parity}

The number of fitness cases for this problem is $2^{4}=16$. For solving the even-4 parity we use a population of 100 individuals evolved for 500 generations. Other TGP parameters are given in Table \ref{tab1}.

\begin{figure}[htbp]
\centerline{\includegraphics{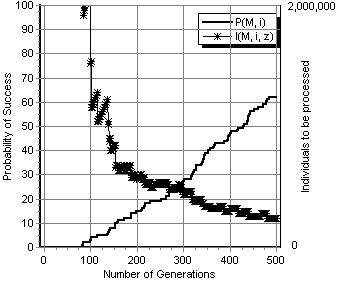}}
\caption{The cumulative probability of success and the computational effort for the even-4 parity problem. Results are averaged over 100 runs.}
\label{fig2}
\end{figure}

The effort and the probability of success of the TGP algorithm are depicted in Figure \ref{fig2}.

The minimum effort is 240,000 and it was obtained at generation 480.

The effort spent by GP for solving the parity problem is 1,276,000. This number was obtained using a population of 4000 individuals \cite{koza1}. One of the solutions evolved by GP has 149 nodes.

If we want to compare the efforts spent by GP and TGP we have to divide the TGP effort (240,000) by 149 (number of nodes in a GP tree). Thus, we obtain the number 1610 which is with almost 3 orders of magnitude better than the result obtained by GP.

\subsection{Even-5 parity}

The number of fitness cases for this problem is $2^{5}=32$. For solving the even-5 parity we use a population of 500 individuals evolved for 1000 generations. Other TGP parameters are given in Table \ref{tab1}.

\begin{figure}[htbp]
\centerline{\includegraphics{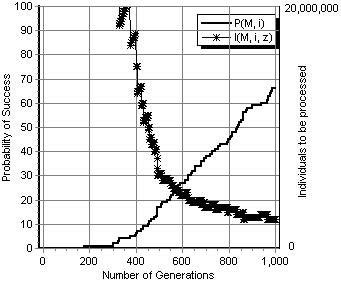}}
\caption{The cumulative probability of success and the computational effort for the even-5 parity problem. Results are averaged over 100 runs.}
\label{fig3}
\end{figure}

The effort and the probability of success of the TGP algorithm are depicted in Figure \ref{fig3}.

The effort is 2,417,500 and it was obtained at generation 967.

For this problem standard GP with a population of 8000 individuals obtained a solution in the $8^{th}$ run \cite{koza1}. No other statistics were given for this problem.

\subsection{Even-6 parity}

The number of fitness cases for this problem is $2^{6}=64$. For solving the even-6 parity we use a population of 1000 individuals evolved for 2500 generations. Other TGP parameters are given in Table \ref{tab1}.

\begin{figure}[htbp]
\centerline{\includegraphics{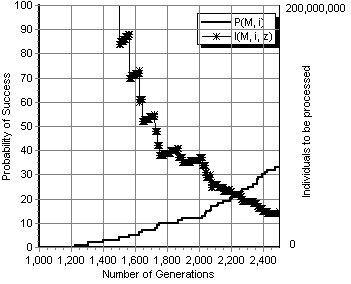}}
\caption{The cumulative probability of success and the computational effort for the even-6 parity problem. Results are averaged over 100 runs.}
\label{fig4}
\end{figure}

The effort and the probability of success of the TGP algorithm are depicted in Figure \ref{fig4}.

The minimum effort is 29,136,000 and it was obtained at generation 2428.

\subsection{Even-7 parity}

The number of fitness cases for this problem is $2^{7}=128$. For solving the even-7 parity we use a population of 2000 individuals evolved for 5000 generations. Other TGP parameters are given in Table \ref{tab1}.

\begin{figure}[htbp]
\centerline{\includegraphics{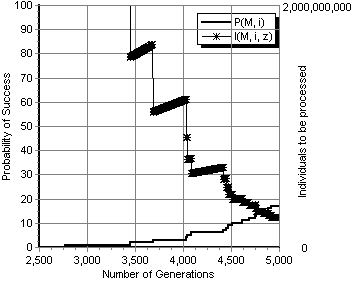}}
\caption{The cumulative probability of success and the computational effort for the even-7 parity problem. Results are averaged over 100 runs.}
\label{fig5}
\end{figure}

The effort and the probability of success of the TGP algorithm are depicted in Figure \ref{fig5}. 

The minimum effort is 245,900,000 and it was obtained at generation 4918.

\subsection{Even-8 parity}

The even-8 parity problem is the most difficult problem analyzed in this paper. The number of fitness cases for this problem is $2^{8}=256$. For solving the even-8 parity we use a population of 5000 individuals evolved for 10000 generations. Other TGP parameters are given in Table \ref{tab1}. Only 10 runs are performed for this problem. In the $9^{th}$ run we obtained a solution.

\section{Summarized results}

Summarized results of applying Traceless Genetic Programming for solving even-parity problems are given in Table \ref{tab2}.

\begin{table}[htbp]
\caption{Summarized results for solving the even-parity problem using TGP. Second column indicates the population size used for solving the problem. The computational effort is given in the $3^{rd}$ column. The numbers in the $4^{th}$ column indicate the generation where the minimum effort was obtained.}
\label{tab2}
\begin{center}
\begin{tabular}
{|p{40pt}|p{40pt}|p{50pt}|p{40pt}|}
\hline
Problem& 
Pop Size& 
Effort& 
Generation \\
\hline
even-3& 
50& 
33,750& 
131 \\
\hline
even-4& 
100& 
240,000& 
480 \\
\hline
even-5& 
500& 
2,417,500& 
967 \\
\hline
even-6& 
1000& 
29,136,000& 
2428 \\
\hline
even-7& 
2000& 
245,900,000& 
4918 \\
\hline
\end{tabular}
\end{center}
\end{table}

Table \ref{tab2} shows that TGP is able to solve the even-parity problems very well. Genetic  Programming without ADF was able to solve instances up to even-5 parity problem within a reasonable time frame and using a reasonable population. Note again that a perfect comparison between GP and TGP cannot be made due to their different individual representation.

Table \ref{tab2} also shows that the effort required for solving the problem increases with one order of magnitude for each instance.

In order to see the effectiveness and the simplicity of the TGP algorithm we give, in Table \ref{tab3}, the time needed for solving these problems using a PIII computer at 850 MHz.

\begin{table}[htbp]
\caption{The average time for obtaining a solution for the even-parity problem using TGP.}
\label{tab3}
\begin{center}
\begin{tabular}
{|p{50pt}|p{50pt}|}
\hline
Problem& 
Time (seconds) \\
\hline
even-3& 
0.2 \\
\hline
even-4& 
0.9 \\
\hline
even-5& 
3.2 \\
\hline
even-6& 
19.3 \\
\hline
even-7& 
92.5 \\
\hline
\end{tabular}
\end{center}
\end{table}

Table \ref{tab3} shows that TGP is very fast. 92 seconds are needed to obtain a solution for the even-7 parity problem. Note that the standard GP was never used for solving the even-7 parity problem due to the expensive computational time.

\section{Limitations of the proposed approach}

There could be a problem with the length of the program evolved by TGP. The number of gates in offspring is the sum of the number of gates in parents + 1. In the case of binary operators the number of gates in a TGP chromosome might increase exponentially.

This problem could be avoided if the selection process takes into account the number of gates of the chosen individuals.

\section{Conclusions}

A new evolutionary technique called Traceless Genetic Programming has been proposed in this paper. TGP uses a new individual representation, new genetic operators and a specific evolutionary algorithm.

TGP has been used for evolving digital circuits for the even parity problems. Numerical experiments have shown that TGP was able to evolve very fast a solution for up to even-8 parity problem. Note that the standard GP evolved (within a reasonable time frame) a solution for up to even-5 parity problem.

\section{Further work}

Further effort will be spent for improving the proposed Traceless Genetic Programming technique. For instance, in \cite{poli1} a Sub-machine code technique was used for improving the performance ot the GP technique. This kind of improvement can be applied for TGP too.

In \cite{poli2} an extended, unbiased set of 16 gates was used for solving the even-parity problems. Numerical experiments shown \cite{poli2} that GP was able to solve up to even-22 parity instance using the considered set of gates. Further numerical experiments with TGP will include the use of the extended set of all possible 16 gates with 2 inputs.

\end{document}